\documentclass{article}
\usepackage{ijcai21}
\usepackage{times}
\usepackage{soul}
\usepackage{url}
\usepackage[hidelinks]{hyperref}
\usepackage[utf8]{inputenc}
\usepackage[small]{caption}
\usepackage{graphicx}
\usepackage{amsmath}
\usepackage{amsthm}
\usepackage{amssymb}
\usepackage{pifont}
\usepackage{booktabs}
\usepackage{algorithm}
\usepackage{algorithmic}
\usepackage{multirow}
\usepackage{array}
\usepackage{enumitem}
\DeclareMathOperator*{\argmin}{arg\,min}
\newcommand{\cmark}{\ding{51}}%
\newcommand{\xmark}{\ding{55}}%
\newcolumntype{P}[1]{>{\centering\arraybackslash}p{#1}}

\title{Automated Machine Learning on Graphs: A Survey}
	\author{
	Ziwei Zhang\footnote{Equal contributions}   \and
	Xin Wang\footnotemark[1] \And
	Wenwu Zhu\footnote{Corresponding author}
	\affiliations
	Tsinghua University, Beijing, China\\
	\emails
	zw-zhang16@mails.tsinghua.edu.cn,
	\{xin\_wang,wwzhu\}@tsinghua.edu.cn
	}

\begin{document}
\maketitle

\begin{abstract}
Machine learning on graphs has been extensively studied	in both academic and industry. However, as the literature on graph learning booms with a vast number of emerging methods and techniques, it becomes increasingly difficult to manually design the optimal machine learning algorithm for different graph-related tasks. To solve this critical challenge, automated machine learning (AutoML) on graphs which combines the strength of graph machine learning and AutoML together, is gaining attention from the research community. Therefore, we comprehensively survey AutoML on graphs in this paper\footnote{We provide a paper collection about AutoML on graphs at \\ \url{https://github.com/THUMNLab/awesome-auto-graph-learning}.}, primarily focusing on hyper-parameter optimization (HPO) and neural architecture search (NAS) for graph machine learning. We further overview libraries related to automated graph machine learning and in-depth discuss AutoGL, the first dedicated open-source library for AutoML on graphs. In the end, we share our insights on future research directions for automated graph machine learning. This paper is the first systematic and comprehensive review of automated machine learning on graphs to the best of our knowledge.

\end{abstract}

\section{Introduction}
Graph data is ubiquitous in our daily life. We can use graphs to model the complex relationships and dependencies between entities ranging from small molecules in proteins and particles in physical simulations to large national-wide power grids and global airlines. Therefore, machine learning on graphs has long been an important research direction for both academics and industry~\cite{newman2018networks}. In particular, network embedding~\cite{cui2018survey,hamilton2017representation,goyal2018graph,cai2018comprehensive} and graph neural networks (GNNs)~\cite{zhang2020deep,wu2020comprehensive,zhou2018graph} have drawn increasing attention in the last decade. They are successfully applied to recommendation systems~\cite{ying2018graph,ma2019learning}, fraud detection~\cite{akoglu2015graph}, bioinformatics~\cite{su2020network,zitnik2017predicting}, physical simulation~\cite{kipf2018neural}, traffic forecasting~\cite{li2018dcrnn_traffic,yu2018spatio}, knowledge representation~\cite{wang2017knowledge}, drug re-purposing~\cite{ioannidis2020few,gysi2020network} and pandemic prediction~\cite{kapoor2020examining} for Covid-19.

Despite the popularity of graph machine learning algorithms, the existing literature heavily relies on manual hyper-parameter or architecture design
to achieve the best performance,
resulting in costly human efforts when a vast number of models emerge for various graph tasks. 
Take GNNs as an example. At least one hundred new general-purpose architectures have been published in top-tier machine learning and data mining conferences in the year 2020 alone, not to mention cross-disciplinary researches of task-specific designs. More and more human efforts are inevitably needed if we stick to the manual try-and-error paradigm in designing the optimal algorithms for targeted tasks.

On the other hand, automated machine learning (AutoML) has been extensively studied to reduce human efforts in developing and deploying machine learning models~\cite{he2020automl,yao2018taking,elshawi2019automated}. Complete AutoML pipelines have the potential to automate every step of machine learning, including auto data collection and cleaning, auto feature engineering, and auto model selection and optimization, etc. Due to the popularity of deep learning models, hyper-parameter optimization (HPO)~\cite{bergstra2012random,bergstra2011algorithms,snoek2012practical} and neural architecture search (NAS)~\cite{elsken2019neural} are most widely studied. AutoML has achieved or surpassed human-level performance~\cite{zoph2017neural,liu2018darts,pham2018efficient} with little human guidance in areas such as computer vision~\cite{zoph2018learning,real2019regularized}.

\textbf{Automated machine learning on graphs}, combining the advantages of AutoML and graph machine learning, naturally serves as a promising research direction to further boost the model performance, which has attracted an increasing number of interests from the community.
In this paper, we provide a comprehensive and systematic review of automated machine learning on graphs, to the best of our knowledge, for the first time. Specifically, we focus on two major topics: HPO and NAS of graph machine learning. For HPO, we focus on how to develop scalable methods. For NAS, we follow the literature and compare different methods from search spaces, search strategies, and performance estimation strategies. How different methods tackle the challenges of AutoML on graphs are discussed along the way. 
Then, we review libraries related to automated graph machine learning and discuss AutoGL, the first dedicated framework and open-source library for automated machine learning on graphs. We highlight the design principles of AutoGL and briefly introduce its usages, which are all specially designed for AutoML on graphs. We believe our review in this paper will significantly facilitate and further promote the studies and applications of automated machine learning on graphs.

The rest of the paper is organized as follows. In Section~\ref{sec:review}, we point out the challenges for automated graph machine learning and briefly introduce basic formulations of machine learning on graphs and AutoML. We comprehensively review HPO on graph machine learning in Section~\ref{sec:hpo} and NAS for graph machine learning in Section~\ref{sec:nas}, followed by our overview and discussions of related libraries in Section~\ref{sec:tool}. Lastly, we outline future research opportunities in Section~\ref{sec:future}.

\section{Automated Machine Learning on Graphs}\label{sec:review}
Automated machine learning on graphs, which non-trivially combines the strength of AutoML and graph machine learning, faces the following challenges.
\begin{itemize}
\item \textbf{The uniqueness of graph machine learning:} Unlike audio, image, or text, which has a grid structure, graph data lies in a non-Euclidean space~\cite{bronstein2017geometric}. Thus, graph machine learning usually has unique architectures and designs. For example, typical NAS methods focus on the search space for convolution and recurrent operations, which is distinct from the building blocks of GNNs~\cite{ijcai2020-195}.
\item \textbf{Complexity and diversity of graph tasks:} As aforementioned, graph tasks per se are complex and diverse, ranging from node-level to graph-level problems, and with different settings, objectives, and constraints~\cite{hu2020open}. How to impose proper \emph{inductive bias} and integrate \emph{domain knowledge} into a graph AutoML method is indispensable.
\item \textbf{Scalability:} Many real graphs such as social networks or the Web are incredibly large-scale with billions of nodes and edges~\cite{zang2018power}. Besides, the nodes in the graph are interconnected and cannot be treated as independent samples. Designing scalable AutoML algorithms for graphs poses significant challenges since both graph machine learning and AutoML are already notorious for being compute-intensive.
\end{itemize}
Approaches with HPO or NAS for graph machine learning reviewed in later sections target handling at least one of these three challenges. We briefly introduce basic problem formulations before moving to the next section.

\subsection{Machine Learning on Graphs}\label{sec:mlgform}
Consider a graph $\mathcal{G} = \left( \mathcal{V},\mathcal{E}\right)$ where $\mathcal{V} = \left\{v_1,v_2,...,v_{\left|\mathcal{V}\right|}\right\}$ is a set of nodes and $\mathcal{E} \subseteq \mathcal{V} \times \mathcal{V}$ is a set of edges. The neighborhood of node $v_i$ is denoted as $\mathcal{N}(i)=  \left\{v_j: (v_i,v_j) \in \mathcal{E}\right\}$. The nodes can also have features denoted as $\mathbf{F}\in \mathbb{R}^{\left|\mathcal{V} \right| \times f}$, where $f$ is the number of features. We use bold uppercases (e.g., $\mathbf{X}$) and bold lowercases (e.g., $\mathbf{x}$) to represent matrices and vectors, respectively.

Most tasks of graph machine learning can be divided into the following two categories: 
\begin{itemize}[leftmargin=0.6cm]
	\item Node-level tasks: the tasks are associated with individual nodes or pairs of nodes. Typical examples include node classification and link prediction. 
	\item Graph-level tasks: the tasks are associated with the whole graph, such as graph classification and graph generation.	
\end{itemize}
For node-level tasks, graph machine learning models usually learn a node representation $\mathbf{H} \in \mathbb{R}^{\left|\mathcal{V}\right| \times d}$ and then adopt a classifier or predictor on the node representation to solve the task. For graph-level tasks, a representation for the whole graph is learned and fed into a classifier/predictor.

GNNs are the current state-of-the-art in learning node and graph representations.
The message-passing framework of GNNs~\cite{gilmer2017neural} is formulated as follows.
\begin{gather}\label{eq:mpnn}
    \mathbf{m}^{(l)}_i = \text{AGG}^{(l)}\left(\left\{a_{ij}^{(l)} \mathbf{W}^{(l)} \mathbf{h}^{(l)}_i, \forall j \in \mathcal{N}(i) \right\} \right) \\
    \mathbf{h}^{(l+1)}_i = \sigma\left(\text{COMBINE}^{(l)}\left[ \mathbf{m}^{(l)}_i, \mathbf{h}^{(l)}_i\right] \right),
\end{gather}
where $\mathbf{h}^{(l)}_i$ denotes the node representation of node $v_i$ in the $l^{th}$ layer, $\mathbf{m}^{(l)}$ is the message for node $v_i$, $\text{AGG}^{(l)}(\cdot)$ is the aggregation function, $a_{ij}^{(l)}$ denotes the weights from node $v_j$ to node $v_i$, $\text{COMBINE}^{(l)}(\cdot)$ is the combining function, $\mathbf{W}^{(l)}$ are learnable weights, and $\sigma(\cdot)$ is an activation function. The node representation is usually initialized as node features $\mathbf{H}^{(0)} = \mathbf{F}$, and the final representation is obtained after $L$ message-passing layers $\mathbf{H} = \mathbf{H}^{(L)}$.

For the graph-level representation, pooling methods (also called readout) are applied to the node representations
\begin{equation}\label{eq:pool}
	\mathbf{h}_{\mathcal{G}} = \text{POOL}\left(\mathbf{H}\right),
\end{equation}
i.e., $\mathbf{h}_{\mathcal{G}}$ is the representation of $\mathcal{G}$.

\subsection{AutoML}\label{sec:automlform}
Many AutoML algorithms such as HPO and NAS can be formulated as the following bi-level optimization problem:
\begin{equation}\label{eq:nasobj}
\begin{split}
	 & \min_{\alpha \in \mathcal{A}} \mathcal{L}_{val}\left(\mathbf{W}^*(\alpha),\alpha\right) \\
      \text{s.t.} \quad & \mathbf{W}^*(\alpha) =  \argmin_{\mathbf{W}} \left(\mathcal{L}_{train}\left(\mathbf{W},\alpha\right)\right) ,
\end{split}
\end{equation}
where $\alpha$ is the optimization objective of the AutoML algorithm, e.g., hyper-parameters in HPO and neural architectures in NAS, $\mathcal{A}$ is the feasible space for the objective, and $\mathbf{W}(\alpha)$ are trainable weights in the graph machine learning models. Essentially, we aim to optimize the objective in the feasible space so that the model achieves the best results in terms of a validation function, and $\mathbf{W}^*$ indicates that the weights are fully optimized in terms of a training function. Different AutoML methods differ in how the feasible space is designed and how the objective functions are instantiated and optimized since directly optimizing Eq.~\eqref{eq:nasobj} requires enumerating and training every feasible objective, which is prohibitive.

Typical formulations of AutoML on graphs need to properly integrate the above formulations in Section~\ref{sec:mlgform} and Section~\ref{sec:automlform} to form a new optimization problem. 

\section{HPO for Graph Machine Learning}\label{sec:hpo}
In this section, we review HPO for machine learning on graphs. The main challenge here is scalability, i.e., a real graph can have billions of nodes and edges, and each trial on the graph is computationally expensive. Next, we elaborate on how different methods tackle the efficiency challenge. Notice that we omit some straightforward HPO methods such as random search and grid search~\cite{bergstra2012random}. 

AutoNE~\cite{tu2019autone} first tackles the efficiency problem of HPO on graphs by proposing a transfer paradigm that samples subgraphs as proxies for the large graph, which is similar in principle to sampling instances in previous HPO methods\cite{hutter2009paramils}. Specifically, AutoNE has three modules: the sampling module, the signature extraction module, and the meta-learning module. In the sampling module, multiple representative subgraphs are sampled from the large graph using a multi-start random walk strategy. Then, AutoNE conducts HPO on the sampled subgraphs using Bayesian optimization~\cite{snoek2012practical} and learns representations of subgraphs using the signature extraction module. Finally, the meta-learning module extracts meta-knowledge from HPO results and representations of subgraphs. AutoNE fine-tunes hyper-parameters on the large graph using the meta-knowledge. In this way, AutoNE achieves satisfactory results while maintaining scalability since multiple HPO trials on the sampled subgraphs and a few HPO trails on the large graph are properly integrated.

JITuNE~\cite{guo2021jitune} proposes to replace the sampling process of AutoNE with graph coarsening to generate a hierarchical graph synopsis. A similarity measurement module is proposed to ensure that the coarsened graph shares sufficient similarity with the large graph. Compared with sampling, such graph synopsis can better preserve graph structural information. Therefore, JITuNE argues that the best hyper-parameters in the graph synopsis can be directly transferred to the large graph. Besides, since the graph synopsis is generated in a hierarchy, the granularity can be more easily adjusted to meet the time constraints of downstream tasks. 

HESGA~\cite{yuan2021novel} proposes another strategy to improve efficiency using a hierarchical evaluation strategy together with evolutionary algorithms. Specifically, HESGA proposes to evaluate the potential of hyper-parameters by interrupting training after a few epochs and calculating the performance gap with respect to the initial performance with random model weights. This gap is used as a fast score to filter out unpromising hyper-parameters. Then, the standard full evaluation, i.e., training until convergence, is adopted as the final assessor to select the best hyper-parameters to be stored in the population of the evolutionary algorithm.

Besides efficiency, AutoGM~\cite{yoon2020autonomous} focuses on proposing a unified framework for various graph machine learning algorithms. Specifically, AutoGM finds that many popular GNNs can be characterized in a framework similar to Eq.~\eqref{eq:mpnn} with five hyper-parameters: the number of message-passing neighbors, the number of message-passing steps, the aggregation function, the dimensionality, and the non-linearity. AutoGM adopts Bayesian optimization to optimize these hyper-parameters.

\section{NAS for Graph Machine Learning}\label{sec:nas}
\begin{table*}[ht]
	\centering
	\begin{scriptsize}
		\begin{tabular}{c|P{0.3cm}P{0.3cm}P{0.4cm}P{0.15cm}P{0.55cm}|P{0.23cm}P{0.43cm}|c|c|c} \toprule
			\multirow{2}{*}{Method}  & \multicolumn{5}{c|}{Search space} & \multicolumn{2}{c|}{Tasks} & \multirow{2}{*}{Search Strategy} & Performance  &\multirow{2}{*}{Other Characteristics}  \\
			                                             & Micro & Macro &Pooling& HP    & Layers & Node   & Graph &                               &Estimation &- \\ \midrule 
			GraphNAS~\shortcite{ijcai2020-195}           &\cmark &\cmark &\xmark &\xmark & Fixed  & \cmark & \xmark& RNN controller + RL           & -         &- \\ 
			AGNN~\shortcite{zhou2019auto}                &\cmark &\xmark &\xmark &\xmark & Fixed  & \cmark & \xmark& Self-designed controller + RL & Inherit weights  &- \\ 
			SNAG~\shortcite{zhao2020simplifying}         &\cmark &\cmark &\xmark &\xmark & Fixed  & \cmark & \xmark& RNN controller + RL           & Inherit weights  &Simplify the micro search space \\ 
			PDNAS~\shortcite{zhao2020probabilistic}      &\cmark &\cmark &\xmark &\xmark & Fixed  & \cmark & \xmark& Differentiable                & Single-path one-shot  &- \\ 
			POSE~\shortcite{ding2020propagation}         &\cmark &\cmark &\xmark &\xmark & Fixed  & \cmark & \xmark& Differentiable                & Single-path one-shot  & Support heterogenous graphs \\ 
			NAS-GNN~\shortcite{nunes2020neural}          &\cmark &\xmark &\xmark &\cmark & Fixed  & \cmark & \xmark& Evolutionary algorithm        & -         &- \\ 
			AutoGraph~\shortcite{li2020autograph}        &\cmark &\cmark &\xmark &\xmark & Various& \cmark & \xmark& Evolutionary algorithm        & -         &- \\ 
			GeneticGNN~\shortcite{shi2020evolutionary}  &\cmark &\xmark &\xmark &\cmark & Fixed  & \cmark & \xmark& Evolutionary algorithm        & -         &- \\ 
			EGAN~\shortcite{zhao2021efficient}           &\cmark &\cmark &\xmark &\xmark & Fixed  & \cmark & \cmark& Differentiable                & One-shot  & Sample small graphs for efficiency \\ 
			NAS-GCN~\shortcite{jiang2020graph}           &\cmark &\cmark &\cmark &\xmark & Fixed  & \xmark & \cmark& Evolutionary algorithm        & -         & Handle edge features               \\ 
			LPGNAS~\shortcite{zhao2020learned}           &\cmark &\cmark &\xmark &\xmark & Fixed  & \cmark & \xmark& Differentiable                & Single-path one-shot  & Search for quantisation options    \\ 
			You~\textit{et al.}~\shortcite{you2020design}&\cmark &\cmark &\xmark &\cmark & Various& \cmark & \cmark & Random search                & -         & Transfer across datasets and tasks \\ 
			SAGS~\shortcite{li2020sgas}                  &\cmark &\xmark &\xmark &\xmark & Fixed  & \cmark & \cmark & Self-designed algorithm      & -         &-    \\ 
			Peng~\textit{et al.}~\shortcite{peng2020learning}&\cmark &\xmark &\xmark &\xmark & Fixed & \xmark & \cmark &CEM-RL~\shortcite{pourchot2018cemrl}&- & Search spatial-temporal modules \\ 
			
			GNAS\shortcite{cai2021rethinking}            &\cmark &\cmark &\xmark &\xmark & Various& \cmark & \cmark & Differentiable & One-shot & - \\
			AutoSTG\shortcite{pan2021autostg}            &\xmark &\cmark &\xmark &\xmark & Fixed & \cmark  & \xmark & Differentiable    & One-shot+meta learning & Search spatial-temporal modules \\
			DSS\shortcite{li2021one} & \cmark & \cmark & \xmark & \xmark & Fixed & \cmark & \xmark & Differentiable & One-shot & Dynamically update search space  \\
			SANE\shortcite{zhao2021search}   & \cmark & \cmark & \xmark &  \xmark & Fixed & \cmark & \xmark & Differentiable & One-shot & - \\	
			AutoAttend\shortcite{guan2021autoattend} & \cmark & \cmark & \xmark & \xmark & Fixed & \cmark & \cmark & Evolutionary algorithm & One-shot & Cross-layer attention \\
			\bottomrule
		\end{tabular}
	\end{scriptsize}
	\caption{A summary of different NAS methods for graph machine learnings.}
	\label{tab:graphnas}
\end{table*}

NAS methods can be compared in three aspects~\cite{elsken2019neural}: search space, search strategy, and performance estimation strategy. Next, we review NAS methods for graph machine learning from these three aspects. We mainly review NAS for GNNs fitting Eq.~\eqref{eq:mpnn} and summarize the characteristics of different methods in Table~\ref{tab:graphnas}. 

\begin{table}
	\centering
	\begin{small}
	\begin{tabular}{c|l} \toprule
		Type        &  Formulation  \\ \midrule
		CONST       & $a_{ij}^{\text{const}} = 1$  \\
		GCN         & $a_{ij}^{\text{gcn}} =  \frac{1}{\sqrt{\left| \mathcal{N}(i)\right| \left| \mathcal{N}(j) \right| }} $ \\
		GAT         & $a_{ij}^{\text{gat}} = \text{LeakyReLU} \left( \text{ATT} \left(\mathbf{W}_a\left[\mathbf{h}_i, \mathbf{h}_j\right]\right) \right)$ \\
		SYM-GAT     & $a_{ij}^{\text{sym}} = a_{ij}^{\text{gat}}+ a_{ji}^{\text{gat}}$ \\
		COS         & $a_{ij}^{\text{cos}} = \text{cos}\left(\mathbf{W}_a \mathbf{h}_i, \mathbf{W}_a \mathbf{h}_j \right)$ \\
		LINEAR      & $a_{ij}^{\text{lin}} = \text{tanh}\left(\text{sum}\left( \mathbf{W}_a \mathbf{h}_i + \mathbf{W}_a \mathbf{h}_j \right) \right)$ \\
		GENE-LINEAR & $a_{ij}^{\text{gene}} = \text{tanh}\left(\text{sum}\left( \mathbf{W}_a \mathbf{h}_i + \mathbf{W}_a \mathbf{h}_j \right) \right)\mathbf{W}_{a}^\prime $ \\ \bottomrule
	\end{tabular}
	\end{small}
	\caption{A typical search space of different aggregation weights.}
	\label{tab:weights}
\end{table}

\subsection{Search Space}\label{sec:searchspace}
The first challenge of NAS on graphs is the search space design since the building blocks of graph machine learning are usually distinct from other deep learning models such as CNNs or RNNs. For GNNs, the search space can be divided into the following four categories.
\subsubsection{Micro Search Space} Following the message-passing framework in Eq.~\eqref{eq:mpnn}, the micro search space defines how nodes exchange messages with others in each layer. Commonly adopted micro search spaces~\cite{ijcai2020-195,zhou2019auto} compose the following components:
\begin{small}
\begin{itemize}[leftmargin = 0.4cm]
	\item Aggregation function $\text{AGG}(\cdot)$: SUM, MEAN, MAX, and MLP.
	\item Aggregation weights $a_{ij}$: common choices are listed in Table~\ref{tab:weights}.
	\item Number of heads when using attentions: 1, 2, 4, 6, 8, 16, etc.
	\item Combining function $\text{COMBINE}(\cdot)$: CONCAT, ADD, and MLP.
	\item Dimensionality of $\mathbf{h}^{l}$: 8, 16, 32, 64, 128, 256, 512, etc.
	\item Non-linear activation function $\sigma(\cdot)$: Sigmoid, Tanh, ReLU, Identity,
	Softplus, Leaky ReLU, ReLU6, and ELU.
\end{itemize}
\end{small}
However, directly searching all these components results in thousands of possible choices in a single message-passing layer. Thus, it may be beneficial to prune the space to focus on a few crucial components depending on applications and domain knowledge~\cite{zhao2020simplifying}.  

\subsubsection{Macro Search Space} Similar to residual connections and dense connections in CNNs, node representations in one layer of GNNs do not necessarily solely depend on the immediate previous layer~\cite{li2019deepgcns,xu2018representation}. These connectivity patterns between layers form the macro search space. Formally, such designs are formulated as
\begin{equation}
	\mathbf{H}^{(l)} = \sum \nolimits_{j <l} \mathcal{F}_{jl} \left(\mathbf{H}^{(j)}\right),
\end{equation}
where $\mathcal{F}_{jl}(\cdot)$ can be the message-passing layer in Eq.~\eqref{eq:mpnn}, ZERO (i.e., not connecting), IDENTITY (e.g., residual connections), or an MLP. Since the dimensionality of $\mathbf{H}^{(j)}$ can vary, IDENTITY can only be adopted if the dimensionality of each layer matches. 

\subsubsection{Pooling Methods} In order to handle graph-level tasks, information from all the nodes is aggregated to form graph-level representations using the pooling operation in Eq~\eqref{eq:pool}. \cite{jiang2020graph} propose a pooling search space including row- or column-wise sum, mean, or maximum, attention pooling, attention sum, and flatten. More advanced methods such as hierarchical pooling~\cite{ying2018hierarchical} could also be added to the search space with careful designs.

\subsubsection{Hyper-parameters} Besides architectures, other training hyper-parameters can be incorporated into the search space, i.e., similar to jointly conducting NAS and HPO. Typical hyper-parameters include the learning rate, the number of epochs, the batch size, the optimizer, the dropout rate, and the regularization strengths such as the weight decay. These hyper-parameters can be jointly optimized with architectures or separately optimized after the best architectures are found. HPO methods in Section~\ref{sec:hpo} can also be combined here.

Another critical choice is the number of message-passing layers. Unlike CNNs, most currently successful GNNs are shallow, e.g., with no more than three layers, possibly due to the over-smoothing problem~\cite{li2018deeper,li2019deepgcns}. Limited by this problem, the existing NAS methods for GNNs usually preset the number of layers as a small fixed number. How to automatically design deep GNNs while integrating techniques to alleviate over-smoothing remains mostly unexplored. On the other hand, NAS may also bring insights to help to tackle the over-smoothing problem~\cite{ijcai2020-195}.

\subsection{Search Strategy}
Search strategies can be broadly divided into three categories: architecture controllers trained with reinforcement learning (RL), differentiable methods, and evolutionary algorithms. 
\subsubsection{Controller + RL} 
A widely adopted NAS search strategy uses a controller to generate the neural architecture descriptions and train the controller with reinforcement learning to maximize the model performance as rewards. For example, if we consider neural architecture descriptions as a sequence, we can use RNNs as the controller~\cite{zoph2017neural}. Such methods can be directly applied to GNNs with a suitable search space and performance evaluation strategy.

\subsubsection{Differentiable} Differentiable NAS methods such as DARTS~\cite{liu2018darts} and SNAS~\cite{xie2019snas} have gained popularity in recent years. Instead of optimizing different operations separately, differentiable methods construct a single super-network (known as the \emph{one-shot model}) containing all possible operations. Formally, we denote
\begin{equation}
	\mathbf{y} = o^{(x,y)}(\mathbf{x}) = \sum \nolimits_{o \in \mathcal{O}} \frac{\exp( \mathbf{z}_{o}^{(x,y)})}{\sum \nolimits_{o^\prime \in \mathcal{O}}\exp( \mathbf{z}_{o^\prime}^{(x,y)})} o(\mathbf{x}), 
\end{equation}
where $o^{(x,y)}(\mathbf{x})$ is an operation in the GNN with input $\mathbf{x}$ and output $\mathbf{y}$, $\mathcal{O}$ are all candidate operations, and $\mathbf{z}^{(x,y)}$ are learnable vectors to control which operation is selected. Briefly speaking, each operation is regarded as a probability distribution of all possible operations.
In this way, the architecture and model weights can be jointly optimized via gradient-based algorithms. The main challenges lie in making the NAS algorithm differentiable, where several techniques such as Gumbel-softmax~\cite{jang2017categorical} and concrete distribution~\cite{maddison2017concrete} are resorted to. When applied to GNNs, slight modification may be needed to incorporate the specific operations defined in the search space.

\subsubsection{Evolutionary Algorithms} Evolutionary algorithms are a class of optimization algorithms inspired by biological evolution. For NAS, randomly generated architectures are considered initial individuals in a population. Then, new architectures are generated using mutations and crossover operations on the population. The architectures are evaluated and selected to form the new population, and the same process is repeated. The best architectures are recorded while updating the population, and the final solutions are obtained after sufficient updating steps.   

For GNNs, regularized evolution (RE) NAS~\cite{real2019regularized} has been widely adopted. RE's core idea is an aging mechanism, i.e., in the selection process, the oldest individuals in the population are removed. 
Genetic-GNN~\cite{shi2020bridging} also proposes an evolution process to alternatively update the GNN architecture and the learning hyper-parameters to find the best fit of each other.

\subsubsection{Combinations} It is also feasible to combine these three types of search strategies mentioned above. For example, AGNN~\cite{zhou2019auto} proposes a reinforced conservative search strategy by adopting both RNNs and evolutionary algorithms in the controller and train the controller with RL. By only generating slightly different architectures, the controller can find well-performing GNNs more efficiently.  \cite{peng2020learning} adopt CEM-RL~\cite{pourchot2018cemrl}, which combines evolutionary and differentiable methods.

\subsection{Performance Estimation Strategy}
Due to the large number of possible architectures, it is infeasible to fully train each architecture independently. Next, we review some performance estimation strategies. 

A commonly adopted ``trick'' to speed up performance estimation is to reduce fidelity~\cite{elsken2019neural}, e.g., by reducing the number of epochs or the number of data points. This strategy can be directly generalized to GNNs.

Another strategy successfully applied to CNNs is sharing weights among different models, known as parameter sharing or weight sharing~\cite{pham2018efficient}. For differentiable NAS with a large one-shot model, parameter sharing is naturally achieved since the architectures and weights are jointly trained. However, training the one-shot model may be difficult since it contains all possible operations. To further speed up the training process, single-path one-shot model~\cite{guo2020single} has been proposed where only one operation between an input and output pair is activated during each pass.

For NAS without a one-shot model, sharing weights among different architecture is more difficult but not entirely impossible. For example, since it is known that some convolutional filters are common feature extractors, inheriting weights from previous architectures is feasible and reasonable in CNNs~\cite{real2017large}. However, since there is still a lack of understandings of what weights in GNNs represent, we need to be more cautious about inheriting weights~\cite{zhao2020simplifying}. AGNN~\cite{zhou2019auto} proposes three constraints for parameter inheritance: same weight shapes, same attention and activation functions, and no parameter sharing in batch normalization and skip connections. 

\subsection{Discussions}
\subsubsection{The Search Space} Besides the basic search space presented in Section~\ref{sec:searchspace}, different graph tasks may require other search spaces. For example,  meta-paths are critical for heterogeneous graphs~\cite{ding2020propagation}, edge features are essential in modeling molecular graphs~\cite{jiang2020graph}, and spatial-temporal modules are needed in skeleton-based recognition~\cite{peng2020learning}. Sampling mechanisms to accelerate GNNs are also critical, especially for large-scale graphs~\cite{ijcai2020-195}. A suitable search space usually requires careful designs and domain knowledge.

\subsubsection{Transferability} It is non-trivial to transfer GNN architectures across different datasets and tasks due to the complexity and diversity of graph tasks. \cite{you2020graph} adopt a fixed set of GNNs as anchors on different tasks and datasets. 
Then, the rank correlation serves as a metric to measure the similarities between different datasets and tasks. The best-performing GNNs of the most similar tasks are transferred to solve the target task. 

\subsubsection{The Efficiency Challenge of Large-scale Graphs} Similar to AutoNE introduced in Section~\ref{sec:hpo}, EGAN~\cite{zhao2021efficient} proposes to sample small graphs as proxies and conduct NAS on the sampled subgraphs to improve the efficiency of NAS. While achieving some progress, more advanced and principle approaches are further needed to handle billion-scale graphs.

\section{Libraries for AutoML on Graphs}\label{sec:tool}

Publicly available libraries are important to facilitate and advance the research and applications of AutoML on graphs. Popular libraries for graph machine learning include PyTorch Geometric \cite{Fey2019fast}, Deep Graph Library \cite{dgl}, GraphNets \cite{graphnet}, AliGraph \cite{aligraph}, Euler \cite{euler}, PBG \cite{pbg}, Stellar Graph \cite{StellarGraph}, Spektral \cite{spektral}, CodDL \cite{cogdl}, OpenNE \cite{openne}, GEM \cite{gem}, Karateclub \cite{karateclub}, DIG \cite{dig}, and classical NetworkX \cite{networkx}. However, these libraries do not support AutoML.

On the other hand, AutoML libraries such as NNI \cite{nni}, AutoKeras \cite{autokeras}, AutoSklearn \cite{autosklearn}, Hyperopt \cite{hyperopt}, TPOT \cite{tpot}, AutoGluon \cite{agtabular}, MLBox \cite{mlbox}, and MLJAR \cite{mljar} are widely adopted. Unfortunately, because of the uniqueness and complexity of graph tasks, they cannot be directly applied to automate graph machine learning.

\begin{figure}[t] 
	\centering
	\includegraphics[width=0.99\linewidth]{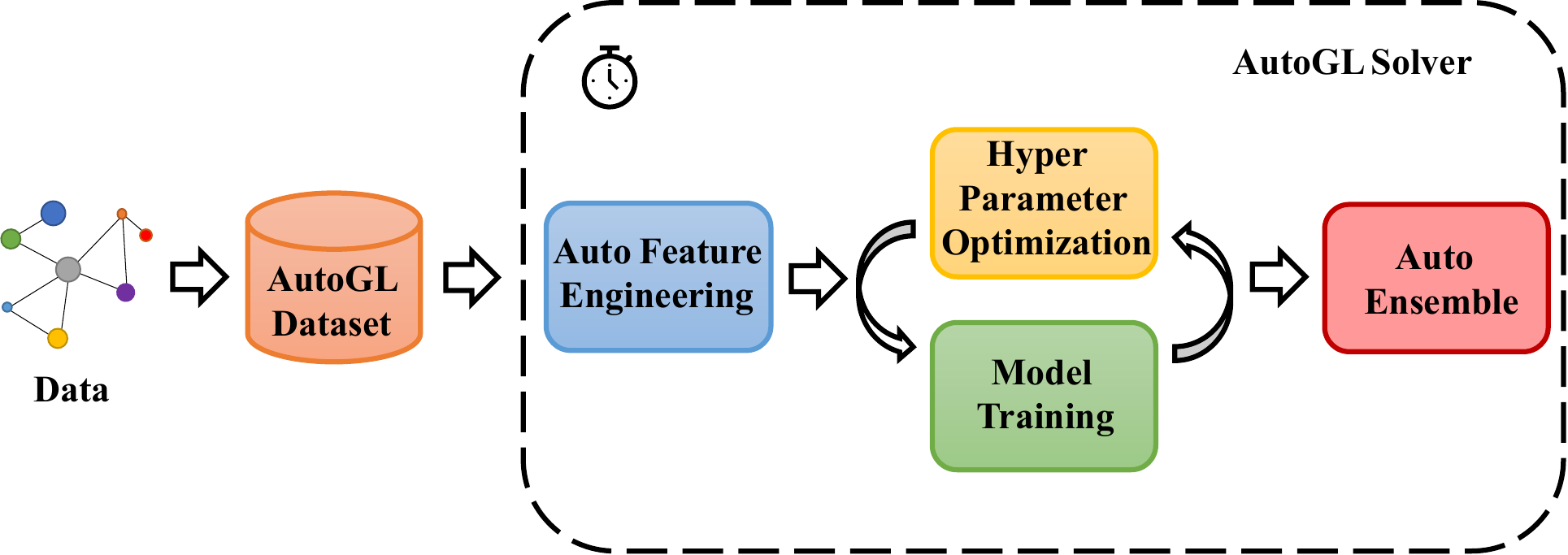} 
	\caption{The overall framework of AutoGL.}
	\label{fig:framework} 
\end{figure}

Recently, some HPO and NAS methods for graphs such as AutoNE \cite{tu2019autone}, AutoGM \cite{yoon2020autonomous}, GraphNAS \cite{ijcai2020-195} GraphGym \cite{you2020design} have open-sourced their codes, facilitating reproducibility and promoting AutoML on graphs. Besides, AutoGL~\cite{guan2021autogl}\footnote{AutoGL homepage: \url{https://mn.cs.tsinghua.edu.cn/AutoGL}}, the first dedicated library for automated graph learning, is developed. Next, we review AutoGL in detail. 
Figure~\ref{fig:framework} shows the overall framework of AutoGL. The main characteristics of AutoGL are three-folded:
\begin{itemize}[leftmargin=0.5cm]
	\item Open-source: all the source codes of AutoGL are publicly available under the MIT license. 
	\item Easy to use: AutoGL is designed to be easy to use. For example, less than ten lines of codes are needed to conduct some quick experiments of AutoGL.  
	\item Flexible to be extended: AutoGL adopts a modular design with high-level base classes API and extensive documentations, allowing flexible and customized extensions.
\end{itemize}
We briefly review the dataset management and four core components of AutoGL: Auto Feature Engineering, Model Training, Hyper-Parameters Optimization, and Auto Ensemble. These components are designed in a modular and object-oriented fashion to enable clear logic flows, easy usages, and flexibility in extending. 

\paragraph{AutoGL Dataset.} inherits from PyTorch Geometric~\cite{Fey2019fast}, covering common benchmarks for node and graph classification, including the recent Open Graph Benchmark~\cite{hu2020open}. Users can easily add customized datasets following documentations.

\paragraph{Auto Feature Engineering.} module first processes the graph data using three categories of operations: generators, where new node and edge features are constructed; selectors, filtering out and compressing useless and meaningless features; sub-graph generators, generating graph-level features. Convenient wrappers are also provided to support PyTorch Geometric and NetworkX~\cite{hagberg2008exploring}. 

\paragraph{Model Training.} module handles the training and evaluation process of graph tasks with two functional sub-modules: model and trainer. Model handles the construction of graph machine learning models, e.g., GNNs, by defining learnable parameters and the forward pass. Trainer controls the optimization process for the given model. Common optimization methods, training controls, and regularization methods are packaged as high-level APIs.

\paragraph{Hyper-Parameter Optimization.} module conducts HPO for a specified model, covering methods presented in Section~\ref{sec:hpo} such as AutoNE and general-purpose algorithms like random search~\cite{bergstra2012random} and Tree Parzen Estimator~\cite{tpe}.
The model training module can specify the hyper-parameters, their types (e.g., integer, numerical, or categorical), and feasible ranges.
Users can also customize HPO algorithms.

\paragraph{Auto Ensemble.} module can automatically integrate the optimized individual models to form a more powerful final model. Two kinds of ensemble methods are provided: voting and stacking. Voting is a simple yet powerful ensemble method that directly averages the output of individual models while stacking trains another meta-model that learns to combine the output of models in a more principled way.
The general linear model (GLM) and gradient boosting machine (GBM) are supported as meta-models. 

\paragraph{AutoGL Solver.} On top of the four modules, another high-level API Solver is proposed to control the overall pipeline. In the Solver, the four modules are organized to form the AutoML solution for graph data. The Solver also provides global controls. For example, the time budget can be explicitly set, and the training/evaluation protocols can be selected.

 AutoGL is still actively updated. Key features to be released shortly include neural architecture search, large-scale datasets support, and more graph tasks. For the most up-to-date information, please visit the project homepage. 
 All kinds of inputs and suggestions are also warmly welcomed.

\section{Future Directions}\label{sec:future}
\begin{itemize}
\item \textbf{Graph models for AutoML}: In this paper, we mainly focus on 
how AutoML methods are extended to graphs. The other direction, i.e., using graphs to help AutoML, is also feasible and promising. For example, we can model neural networks as a directed acyclic graph (DAG) to analyze their structures~\cite{xie2019exploring,you2020graph} or adopt GNNs to facilitate NAS~\cite{zhang2018graph,dudziak2020brp,shi2020bridging}. Ultimately, we expect graphs and AutoML to form tighter connections and further facilitate each other.
\item \textbf{Robustness and explainability}: Since many graph applications are risk-sensitive, e.g., finance and healthcare, model robustness and explainability are indispensable for actual usages. Though there exist some initial studies on the robustness~\cite{sun2018adversarial} and explainability~\cite{yuan2020explainability} of graph machine learning, how to generalize these techniques into AutoML on graphs remains to be further explored \cite{wang2021towards}.
\item \textbf{Hardware-aware models}: To further improve the scalability of automated machine learning on graphs, hardware-aware models may be a critical step, especially in real industrial environments. Both hardware-aware graph models~\cite{auten2020hardware} and hardware-aware AutoML models~\cite{cai2018proxylessnas,tan2019mnasnet,jiang2020hardware} have been studied, but integrating these techniques still poses significant challenges.
\item \textbf{Comprehensive evaluation protocols}: Currently, most AutoML on graphs are tested on small traditional benchmarks such as three citation graphs, i.e., Cora, CiteSeer, and PubMed~\cite{sen2008collective}. However, these benchmarks have been identified as insufficient to compare different graph machine learning models~\cite{shchur2018pitfalls}, not to mention AutoML on graphs. More comprehensive evaluation protocols are needed, e.g., on recently proposed graph machine learning benchmarks~\cite{hu2020open,dwivedi2020benchmarking} or new dedicated graph AutoML benchmarks similar to the NAS-bench series~\cite{ying2019bench}.
\end{itemize}

\section*{Acknowledgments}
This work was supported in part by the National Key Research and Development Program of China (No.2020AAA0106300, 2020AAA0107800,  2018AAA0102000) and National Natural Science Foundation of China No.62050110.

\bibliographystyle{unsrt}
\bibliography{ijcai21}

\end{document}